\documentclass{itatnew}

\makeatletter
\def\blfootnote{\xdef\@thefnmark{}\@footnotetext}
\makeatother

\begin{document}

\title{Video Scene Location Recognition with Neural Networks}

\author{Lukáš Korel\inst{1} \and Petr Pulc\inst{1} \and Jiří Tumpach\inst{2} \and Martin Holeňa\inst{3}}

\institute{Faculty of Information Technology, Czech Technical University, Prague, Czech Republic
\and
Faculty of Mathematics and Physics, Charles University, Prague, Czech Republic
\and
Institute of Computer Science, Academy of Sciences of the Czech Republic, Prague, Czech Republic}

\maketitle              
\blfootnote{Copyright \copyright 2021 for this paper by its authors. Use permitted under Creative Commons License Attribution 4.0 International (CC BY 4.0).}

\begin{abstract}
This paper provides an insight into the possibility of scene recognition from a video sequence with a small set of repeated shooting locations (such as in television series) using artificial neural networks. The basic idea of the presented approach is to select a set of frames from each scene, transform them by a pre-trained single-image pre-processing convolutional network, and classify the scene location with subsequent layers of the neural network. The considered networks have been tested and compared on a dataset obtained from The Big Bang Theory television series. We have investigated different neural network layers to combine individual frames, particularly AveragePooling, MaxPooling, Product, Flatten, LSTM, and Bidirectional LSTM layers. We have observed that only some of the approaches are suitable for the task at hand.
\end{abstract}
\section{Introduction}
People watching videos are able to recognize where the current scene is located. When watching some film or serial, they are able to recognize that a new scene is on the same place they have already seen. Finally, people are able to understand scenes hierarchy. All this supports human comprehensibility of videos.

The role of location identification in scene recognition by humans motivated our research into scene location classification by artificial neural networks (ANNs). A more ambitious goal would be a make system able to remember unknown video locations and using this data identify video scene that is located in that location and mark it with the same label. This paper reports a work in progress in that direction. It describes the employed methodology and presents first experimental results obtained with six kinds of neural networks.

The rest of the paper is organized as follows. The next section is about existing approaches to solve this problem. Section 3 is divided to two parts. The first one is about data preparation before their usage in ANNs. The second one is about design of the ANNs in our experiments. Finally, Section 4 -- the last section before the conclusion shows our results of experiments with these ANNs.

\section{ANN-Based Scene Classification}
The problem of scene classification has been studied for many years. There are many approaches based on neural networks, where an ANN using huge amount of images learned to recognize the type of given scene (for example, a kitchen, a bedroom, etc.). For this case several datasets are available. One example is \cite{DatasetPlaces}, but it does not specify locations, so this and similar datasets are not usable for our task.

However, our classification problem is different. We want to train an ANN able to recognize a particular location (for example “Springfield-EverGreenTerrace-742-floor2-bathroom”), which can me recorded by camera from many angles (typically, some object can be occluded by other objects from some angles).

One approach using ANN to solve this task is described in \cite{MSRCNN}, there convolutional networks were used. The difference to our approach is on the one hand in the extraction and usage of video images, on the other hand in types of ANN layers.

Another approach is described in \cite{HLIRSC&SFS}. The authors propose a high-level image representation, called Object Bank, where an image is represented as a scale-invariant response map of a large number of pre-trained generic object detectors. Leveraging on the Object Bank representation, good performances on high level visual recognition tasks can be achieved with simple off-the-shelf classifiers such as logistic regression and linear SVM.

\section{Methodology}

\subsection{Data Preparation}
Video data consists of large video files. Therefore, the first task of video data preparation consists in loading the data that is currently needed.

We have evaluated the distribution of the data used for ANN training. We have found there are some scenes with low occurence, whereas others occur up to 30 times more frequently compared to them. Hence, the second task of video data preparation is to increase the uniformity of their distribution, to prevent biasing the ANN to most frequent classes. This is achieved due to undersampling the frequent classes in the training data.

\begin{figure}[hbtp]
  \centering
    \includegraphics[width=0.48\textwidth]{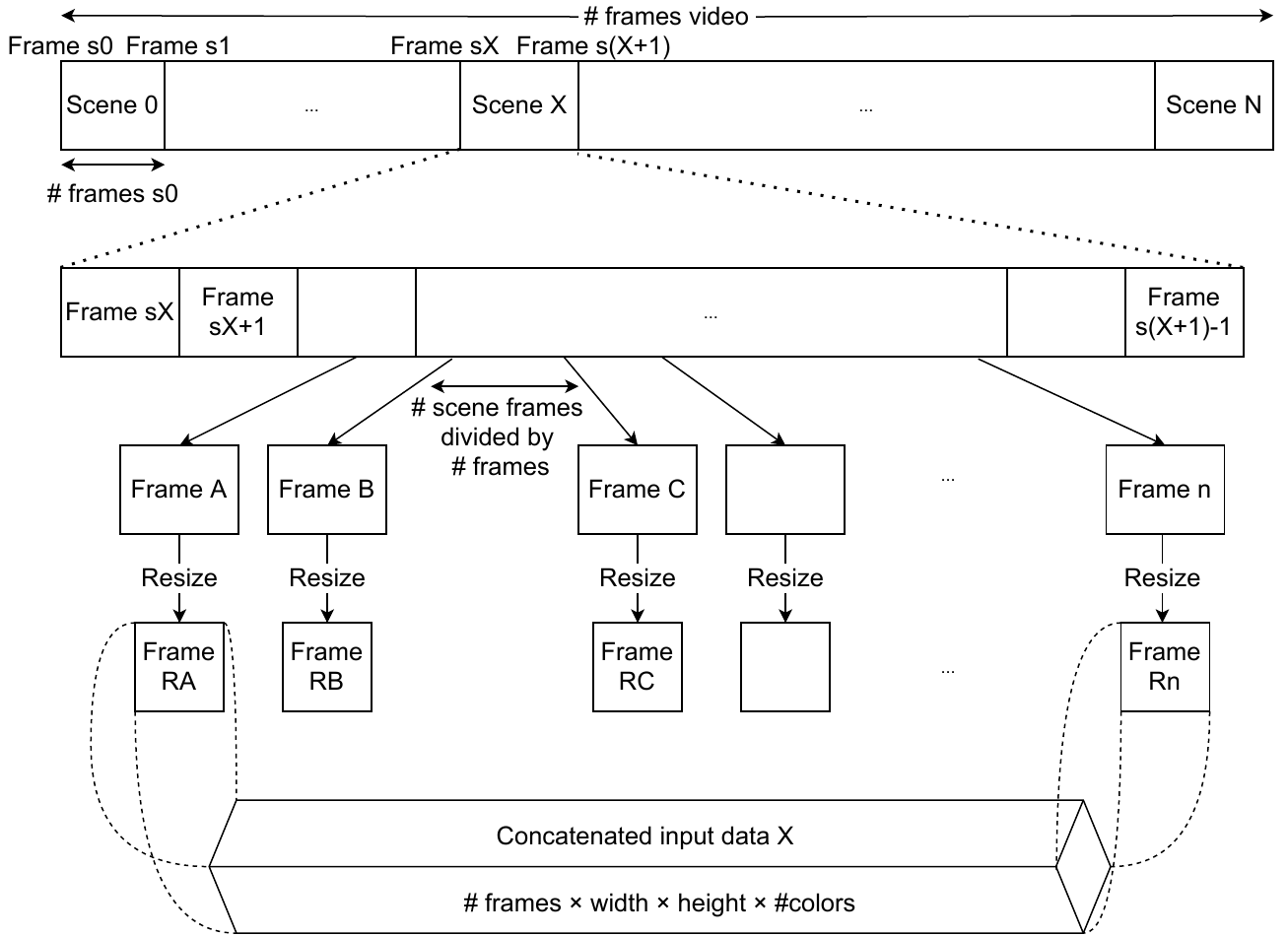}
    \caption{Input preparation for a neural network}
    \label{inputPrep}
\end{figure}

The input consists of video files and a text file. The video files are divided into independent episodes. The text file is contains manually created metainformation about every scene. Every row contains metainformation about one scene. The scene is understand as sequence of frames, that are not interrupted by another frame with different scene location label. Every row contains a relative path to the source video file, the frame number where the scene begins and the count of the its frames. Figure \ref{inputPrep} outlines how frames are extracted and prepared for an ANNs. For ANNs training, we select from each target scene a constant count 20 frames (denoted \# frames in Figure \ref{inputPrep}). To get most informative representation of the considered scene, frames for sampling are taken from the whole length of the scene. This, in particular, prevents to select frames only within a short time interval. Each scene has its own frame distance computed from its frames count: $$SL = \frac{SF}{F}$$ where SF is the count of scene frames, F is the considered constant count of selected frames and SL is the distance between two selected frames in the scene. After frames extraction, every frame is reshaped to an input 3D matrix for the ANN. Finally the reshaped frames are merged to one input matrix for the neural network.

\subsection{Used Neural Networks and Their Design}
Our first idea was to create a complex neural network based on different layers. However, there were too many parameters to train in view of the amount of data that we had. Therefore, we have decided to use transfer learning from some pretrained network.

Because our data are actually images, we considered only ANNs pretrained on image datasets in particular ResNet50 \cite{ResNet}, ResNet101 \cite{ResNet} and VGGnet \cite{VGGnet}. Finally, we have decided to use VGGnet due to its small size.

\begin{figure}[hbtp]
  \centering
    \includegraphics[width=0.29\textwidth]{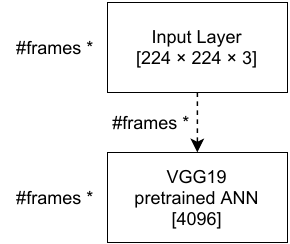}
    \caption{First, untrainable part of our neural network, where the Input Layer represents frame with resolution $224 \times 224$ in BGR colors and output is vector with length 4096, which is output from VGG19 network without last two layers}
    \label{NNbase}
\end{figure}

Hence, ANNs which we trained on our data are composed of two parts. The first part, depicted in Figure \ref{NNbase} is based on the VGGnet. At the input, we have 20 frames (resolution $224 \times 224$, BGR colors) from one scene. This is processed by a pretrained VGG19 neural network without two top layers. The two top layers were removed due to transfer learning. Its output is a vector with size $4096$. For the 20 input frames we have 20 vectors with size 4096. These vectors are merged to a 2D matrix with size $20 \times 4096$.

For the second part, forming the upper layers of the final network, we have considered six possibilities: a product layer, a flatten layer, an average pooling layer, a max pooling layer, an LSTM layer and a bidirectional LSTM layer. All of them, as well as the VCGnet, will be described below. Each of listed layers is preceded by a Dense layer. The Dense layer returns matrix $20 \times 12$, where number 12 is equal to the number of classes. With this output every model works differently.

\paragraph{VGGnet}
The VGGNets \cite{VGGnet} were originally developed for object recognition and detection. They have deep convolutional architectures with smaller sizes of convolutional kernel $(3 \times 3)$, stride $(1 \times 1)$, and pooling window $(2 \times 2)$. There are different network structures, ranging from 11 layers to 19 layers. The model capability is increased when the network is deeper, but imposing a heavier computational cost.

\begin{figure*}[hbtp]
  \centering
    \includegraphics[width=0.9\textwidth]{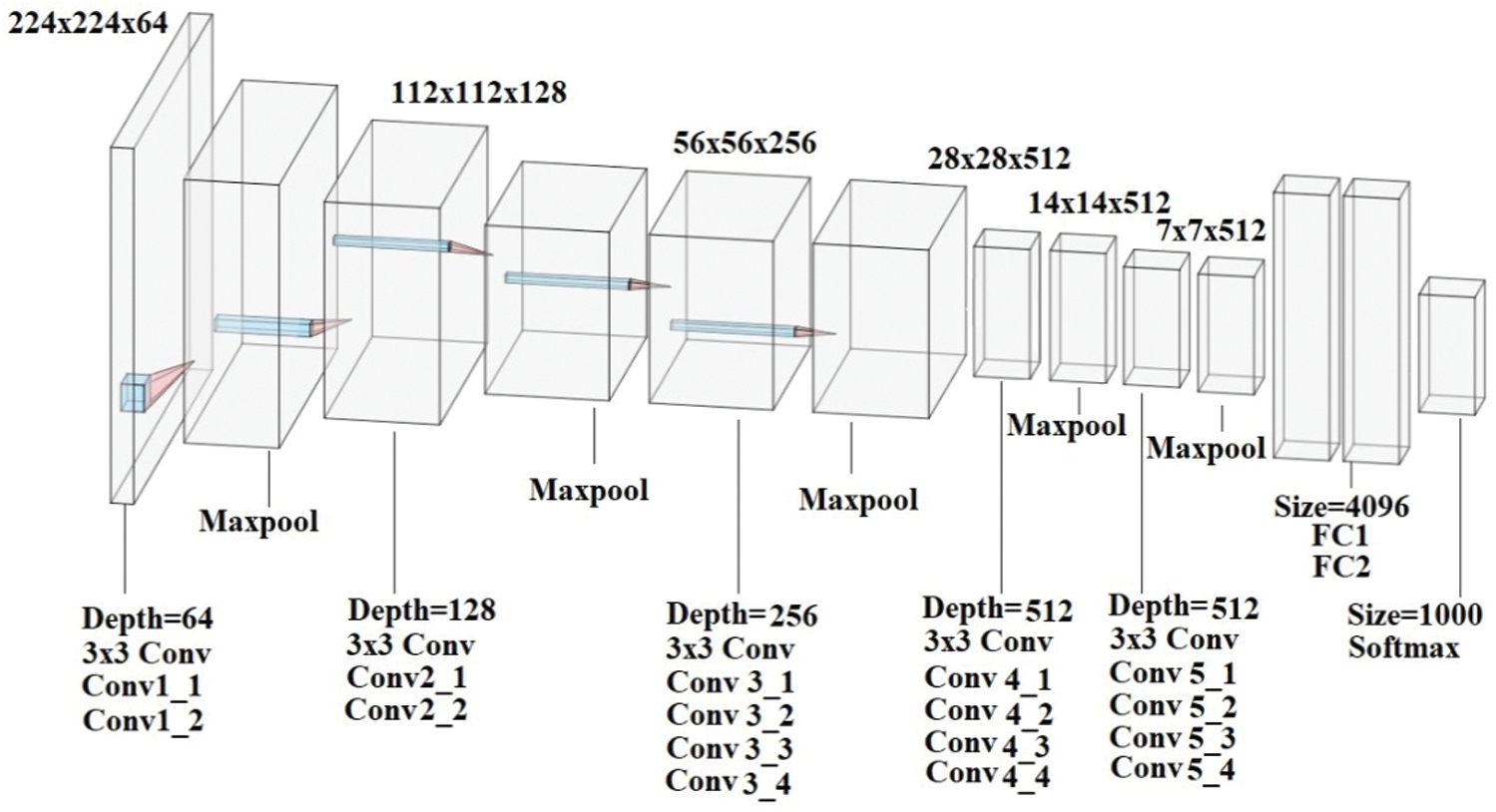}
    \caption{Architecture of the used VGG19 model \cite{VGG19arch}, in our network is used without FC1, FC2 and Softmax layers}
    \label{VGG19img}
\end{figure*}

We have used the VGG19 model (VGG network with 19 layers) from the Keras library in our case. This model \cite{ILSVRC} won the 1st and 2nd place in the 2014 ImageNet Large Scale Visual Recognition Challenge in the 2 categories called \textbf{object localization} and \textbf{image classification}, respectively. It achieves 92.7\% in image classification on Caltech-101, top-5 test accuracy on ImageNet dataset which contains 14 million images belonging to 1000 classes. The architecture of the VGG19 model is depicted in figure \ref{VGG19img}.

\subsubsection{Product array}
In this approach, we apply a product array layer to all output vectors from the dense layer. A Product array layer computes product of all values in chosen dimension of an n-dimensional array and returns an n-1-dimensional array.

\begin{figure}[hbtp]
  \centering
    \includegraphics[width=0.19\textwidth]{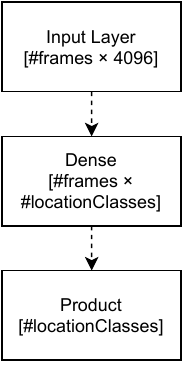}
    \caption{Trainable part of the neural network based on a product layer}
    \label{NNproduct}
\end{figure}

A model with a product layer is outlined in Figure \ref{NNproduct}. The output from a Product layer is  one number for each class, i.e. scene location, so our result is vector with $12$ numbers. It returns a probability distribution over the set of scene locations.

\subsubsection{Flatten}
In this approach, we apply a flatten layer to all output vectors from the dense layer. A Flatten layer creates one long vector from matrix so, that all rows are in sequence.

\begin{figure}[hbtp]
  \centering
    \includegraphics[width=0.19\textwidth]{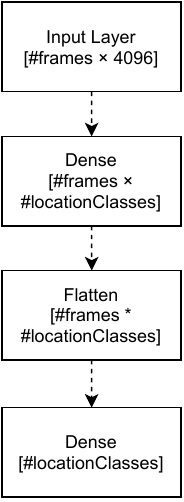}
    \caption{Trainable part of the neural network based on a flatten layer}
    \label{NNflatten}
\end{figure}

A model with a flatten layer is outlined in Figure \ref{NNflatten}. After the input and a dense layer, a flatten layer follows, which returns long vector with $12 * 20$ numbers in this case. It is followed by a second dense layer. Its output has again a dimension equal to the number of classes and it returns a probability distribution over the set of scene locations.

\subsubsection{Average Pooling}
\begin{figure}[hbtp]
  \centering
    \includegraphics[width=0.19\textwidth]{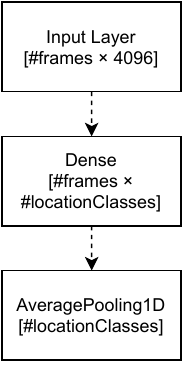}
    \caption{Trainable part of the neural network based on an average-pooling layer}
    \label{NNavg}
\end{figure}

In this approach, we apply average pooling to all output vectors from the dense layer part of the network (Figure \ref{NNavg}). An average-pooling layer computes the average of values assigned to subsets of its preceding layer that are such that:
\begin{itemize}
\item they partition the preceding layer, i.e., that layer equals their union and they are mutually disjoint;
\item they are identically sized.
\end{itemize}
Taking into account these two conditions, the size $p_1\times\dots\times p_D$ of the preceding layer  and the size $r_1\times \dots \times r_D$ of the sets forming its partition determine the size of the average-pooling layer.

In this case, an Average Pooling layer's forming sets size is $20 \times 1$. Using this size in average-pooling layer, we get again one number for each class, which returns a probability distribution over the set of scene locations.

Apart form average pooling, we have tried also max pooling. However, it led to substantially worse results. Its classification of the scene location was typically based on people or items in the foreground, not on the scene as a whole.

Although using the average-pooling layer is easy, it gives acceptable results. The number of trainable parameters of the network is then low, which makes it suitable for our comparatively small dataset.

\subsubsection{Long Short Term Memory}
An LSTM layer is used for classification of sequences of feature vectors, or equivalently, multidimensional time series with discrete time. Alternatively, that layer can be also employed to obtain sequences of such classifications, i.e., in situations when the neural network input is a sequence of feature vectors and its output is a a sequence of classes, in our case of scene locations. LSTM layers are intended for recurrent signal propagation, and differently to other commonly encountered layers, they consists not of simple neurons, but of units with their own inner structure. Several variants of such a structure have been proposed (e.g.,~\cite{gers99learning,graves12supervised}), but all of them include at least the following four components:
\begin{itemize}
\item \emph{Memory cells} can store values, aka cell states, for an arbitrary time. They have no activation function, thus their output is actually a biased linear combination of unit inputs and of the values coming through recurrent connections.  
\item \emph{Input gate} controls the extent to which values from the previous unit within the layer or from the preceding layer influence the value stored in the memory cell. It has a sigmoidal activation function, which is applied to a biased linear combination of the input and recurrent connections, though its bias and synaptic weights are specific and in general different from the bias and synaptic weights of the memory cell.  
\item \emph{Forget gate} controls the extent to which the memory cell state is suppressed. It again has a sigmoidal activation function, which is applied to a specific biased linear combination of input and recurrent connections.
\item \emph{Output gate} controls the extent to which the memory cell state influences the unit output. Also this gate has a sigmoidal activation function, which is applied to a specific biased linear combination of input and recurrent connections, and subsequently composed either directly with the cell state or with its sigmoidal transformation, using a different sigmoid than is used by the gates. 
\end{itemize}   

Hence using LSTM layers a more sophisticated approach compared to simple average pooling. A LSTM, layer can keep hidden state through time with information about previous frames.

\begin{figure}[hbtp]
  \centering
    \includegraphics[width=0.19\textwidth]{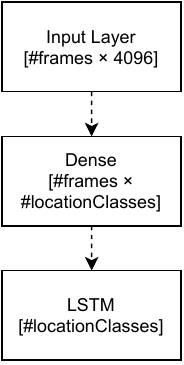}
    \caption{Trainable part of the neural network based on an LSTM layer}
    \label{NNlstm}
\end{figure}

Figure \ref{NNlstm} shows that the input to an LSTM layer is a 2D matrix. Its rows are ordered by the time of frames from the input scene. Every input frame in the network is represented by one vector. The output from the LSTM layer is a vector of the same size as in previous approaches, which returns a probability distribution over the set of scene locations.

\subsubsection{Bidirectional Long Short Term Memory}
An LSTM, due to its hidden state, preserves information from inputs that has already passed through it. Unidirectional LSTM only preserves information from the past because the only inputs it has seen are from the past. A Bidirectional LSTM runs inputs in two ways, one from the past to the future and one from the future to the past. To this end, it combines two hidden states, one for each direction.

\begin{figure}[hbtp]
  \centering
    \includegraphics[width=0.19\textwidth]{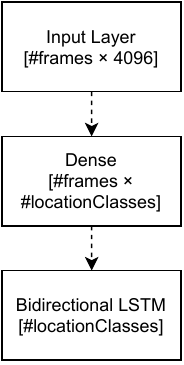}
    \caption{Trainable part of the neural network based on a bidirectional LSTM layer}
    \label{NNblstm}
\end{figure}

Figure \ref{NNblstm} shows that the input to a bidirectional LSTM layer is the same as the input to a LSTM layer. Every input frame in the network is represented by one vector. The output from the Bidirectional LSTM layer is a vector of the same size as in previous approaches, which returns a probability distribution over the set of scene locations.

\section{Experiments}

\subsection{Experimental Setup}
The ANNs for scene location classification were implemented in the libraries Python language using TensorFlow and Keras. Neural network training was accelerated using a NVIDIA GPU. The versions of the employed hardware and software are listed in Table \ref{setup}. For image preparation, OpenCV and Numpy were used. The routine for preparing frames is a generator. It has lower capacity requirements, because data are loaded just in time when they are needed and memory is released after the data have been used for ANN. All non-image information about inputs (video location, scenes information, etc.) are processed in text format by Pandas.

\begin{table}[hbtp]
	\caption{Versions of the employed hardware and software}
	\label{setup}
	\begin{center}
	\begin{tabular}{|l|l|}
	\hline
	CPU cores & 2 \\ \hline
	GPU compute capability & 3.5 and higher \\ \hline
	OS & Linux 5.4.0 \\ \hline
	CUDA & 11.3 \\ \hline
	Python & 3.8.6 \\ \hline
	TensorFlow & 2.3.1 \\ \hline
	Keras & 2.4.0 \\ \hline
	OpenCV & 4.5.2 \\ \hline
	\end{tabular}
	\end{center}
\end{table}

We have 17 independent datasets prepared by ourselves from proprietary videos of the The Big Bang Theory series, thus the datasets can't be public. Each dataset originates from one episode of the series. Each experiment was trained with one dataset, so results are independent as well. So we can compare behavior of the models with different datasets.

Our algorithm to select data in training routine is based on oversampling. It randomly selects target class and from the whole training dataset is randomly select source scene with replacement. This algorithm is applied due to an unbalanced proportion of different target classes. Thanks to this method, all targets are distributed equally and the network does not overfit a highly represented class.

\subsection{Results}
The differences between the models considered in the second, trained part of the network were tested for significance by the Friedman test. The basic null hypotheses that the                   mean classification accuracy for all 6 models coincides was strongly rejected,  with the achieved significance $p = 2.8 \times 10^{-13}$. For the post-hoc analysis, we employed the Wilcoxon signed rank test with two-sided alternative for all 15 pairs of theconsidered models, because of the inconsistence of  the more commonly used mean ranks post-hoc test, to which recently Benavoli et~al.  pointed out \cite{benavoli16should}. For correction to multiple hypotheses testing, we used the Holm method \cite{garcia08extension}. The results are included the comparison between models in Table \ref{tests}.

\begin{table*}[hbtp]
	\caption{Comparison of accuracy results on all 17 episode datasets. The values in the table are counts of datasets, in which the model in row has higher accuracy compared to the model in column. If the difference is not significant in the Wilcoxon test than the count is in italic. If the difference is significant, then the higher count is in bold.}
	\label{tests}
	\begin{center}
	\begin{tabular}{lrrrrrrr}
\hline
{} & Product & Flatten & Average &  Max & LSTM & BidirectionalLSTM &  SummaryScore \\
\hline
Product & X &  \textbf{16} &   \textit{6} &  \textbf{16} &   \textit{5} &   1 &  44 \\
Flatten &   1 & X & 0 &  \textit{10} &   0 &   0 &  11 \\
Average &  \textit{11} &  \textbf{17} & X &  \textbf{17} &   3 &   1 &  49 \\
Max &   1 &   \textit{6} &   0 & X &   0 &   0 &   7 \\
LSTM &  \textit{12} &  \textbf{17} &  14 &  \textbf{17} & X &   3 &  63 \\
BidirectionalLSTM &  \textbf{16} &  \textbf{17} &  \textbf{15} &  \textbf{17} &  \textit{14} & X &  79 \\
\hline
	\end{tabular}
	\end{center}
\end{table*}

Summary statistics of the predictive accuracy of classification all 17 episode datasets are in Table \ref{stats}. Every experiment was performed on every dataset at least 7 times. The table is complemented with results for individual episodes, depicted in box plots.

\begin{figure*}[hbtp]
  \centering
    \includegraphics[width=0.99\textwidth]{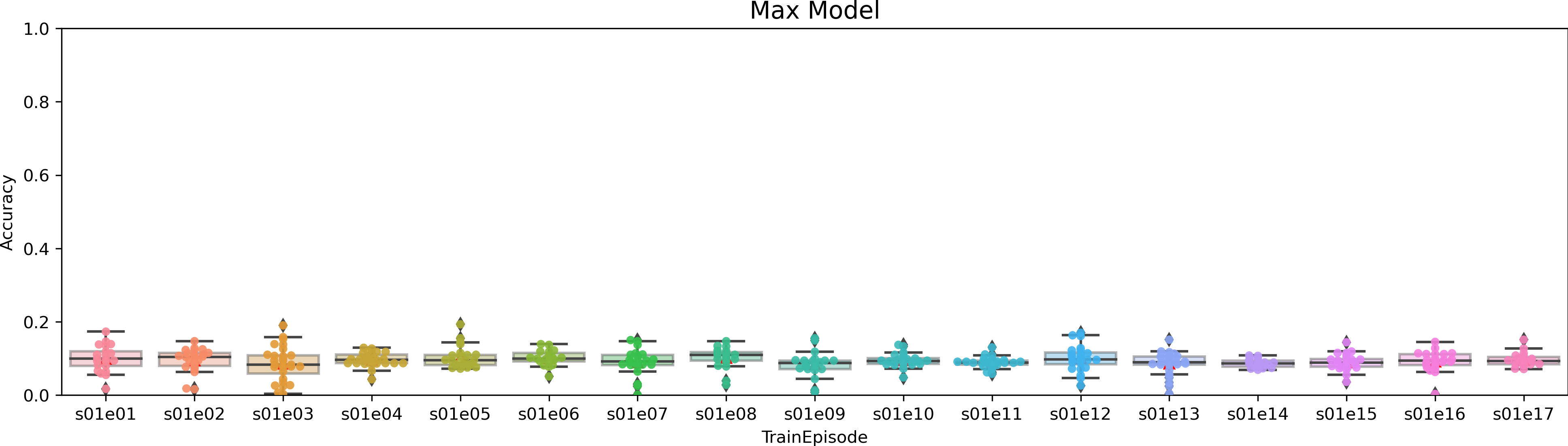}
    \caption{Box plot with results obtained using the max-pooling model}
    \label{BoxMax}
\end{figure*}

The model with a max-pooling layer had the worst results (Figure \ref{BoxMax}) of all experiments. Its overall mean accuracy was around 10 \%. This is only slighty higher than random choice which is $1/12$. The model was not able to achieve better accuracy than 20 \%. Its results were stable and standard deviation was very low.

\begin{figure*}[hbtp]
  \centering
    \includegraphics[width=0.99\textwidth]{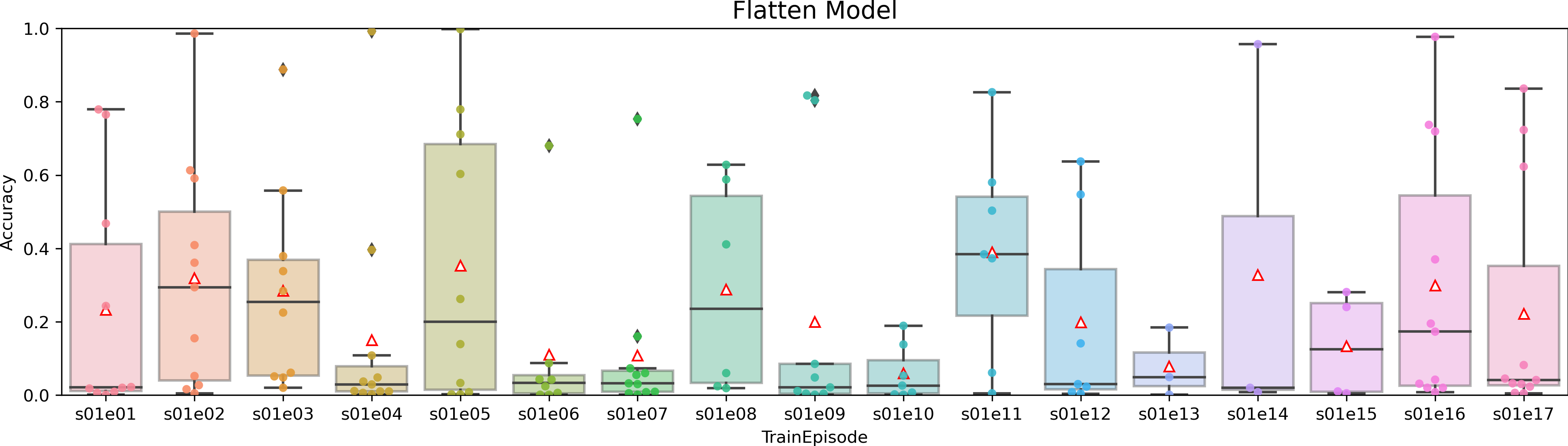}
    \caption{Box plot with results obtained using the flatten model}
    \label{BoxFlatten}
\end{figure*}

Slightly better results (Figure \ref{BoxFlatten}) had the model with the a flatten layer, it was sometimes able to achieve a high accuracy, but its standard deviation was very high. On the other hand, results for some other episodes were not better than those of the max-pooling model.

\begin{figure*}[hbtp]
  \centering
    \includegraphics[width=0.99\textwidth]{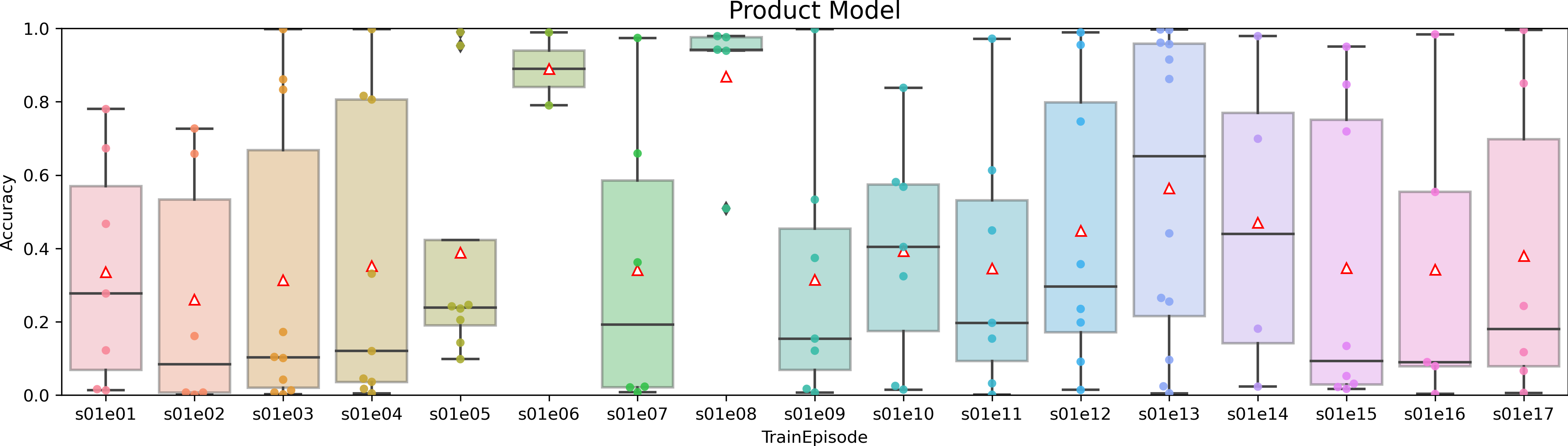}
    \caption{Box plot with results obtained using the product model}
    \label{BoxProduct}
\end{figure*}

A better solution is the product model, whose predictive accuracy (Figure \ref{BoxProduct}) was for several episodes higher than 80 \%. On the other hand, other episodes had only slightly better results than the flatten model. And it had the highest standard deviation among all considered models.

\begin{figure*}[hbtp]
  \centering
    \includegraphics[width=0.99\textwidth]{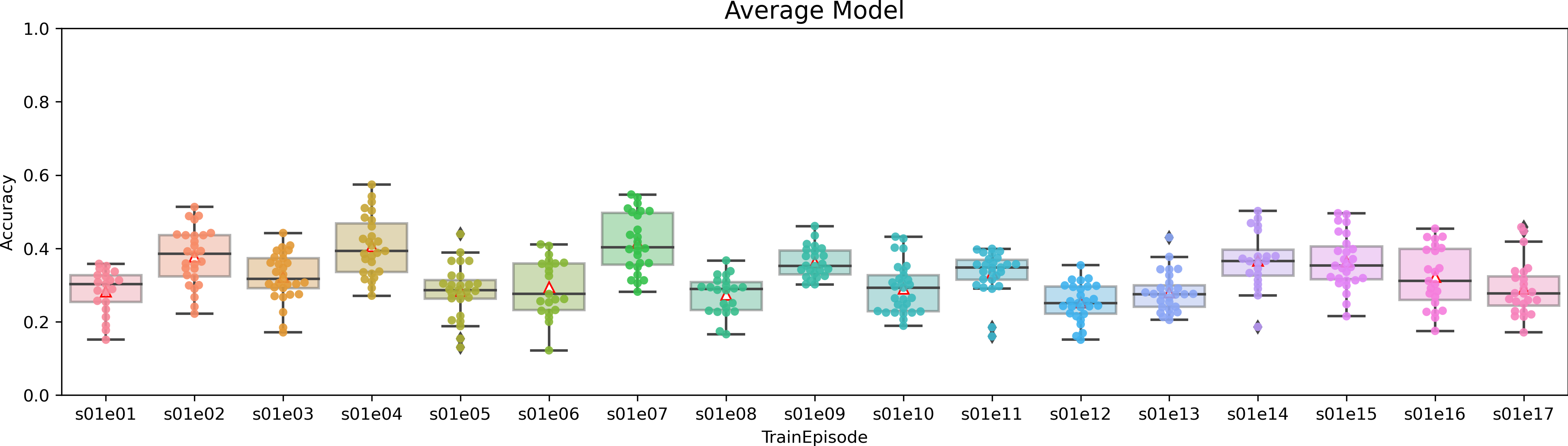}
    \caption{Box plot with results obtained using the average-pooling model}
    \label{BoxAverage}
\end{figure*}

The most stable results (Figure \ref{BoxAverage}) with good accuracy had the model based on average-pooling layer. Its mean accuracy was 32 \% and for no episode, the accuracy was substantially different.

\begin{figure*}[hbtp]
  \centering
    \includegraphics[width=0.99\textwidth]{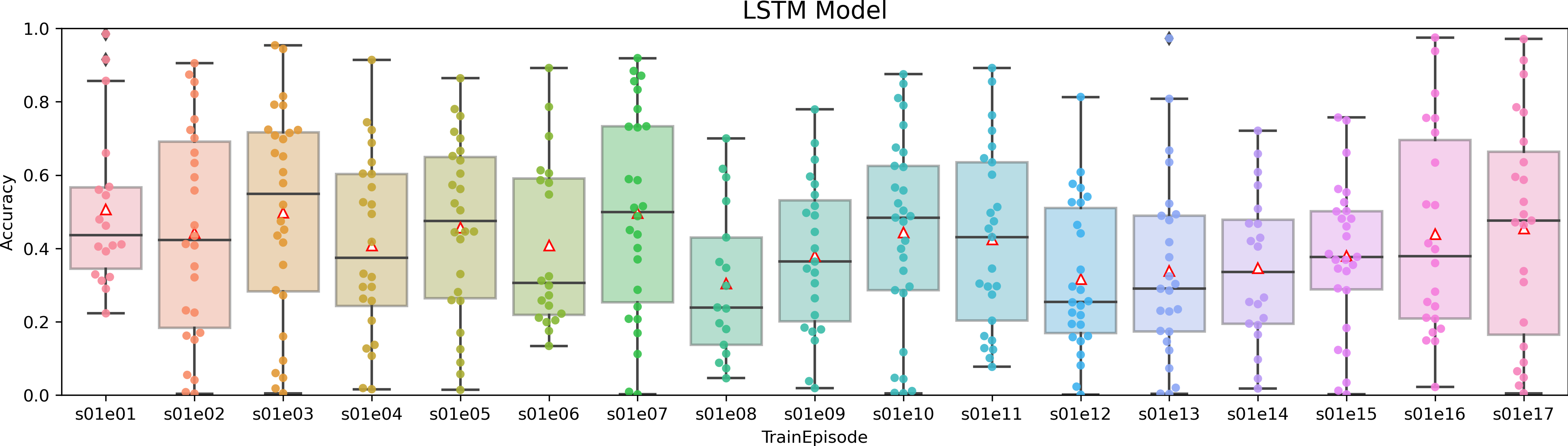}
    \caption{Box plot with results obtained using the LSTM model}
    \label{BoxLSTM}
\end{figure*}

The model with unidirectional LSTM layer had the second mean accuracy of considered our models (Figure \ref{BoxLSTM}). Its internal memory brings advantage in compare over the previous approaches, over 40 \%, though also a comparatively high standard deviation.

\begin{figure*}[hbtp]
  \centering
    \includegraphics[width=0.99\textwidth]{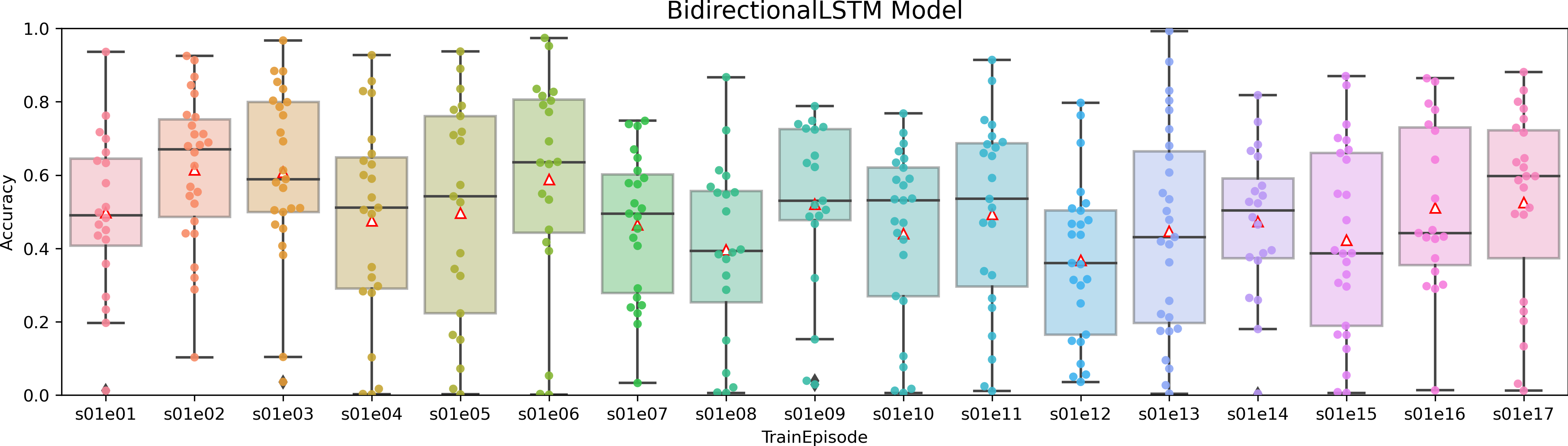}
    \caption{Box plot with results obtained using the bidirectional LSTM model}
    \label{BoxBidirectional}
\end{figure*}

The highest mean accuracy had the model with a bidirectional LSTM layer (Figure \ref{BoxBidirectional}). It had a similar standard deviation as the one with a unidirectional LSTM, but an accuracy mean nearly 50 \%.

\begin{table*}[hbtp]
	\caption{Aggregated predictive accuracy over all 17 datasets [\%]}
	\label{stats}
	\begin{center}
	\begin{tabular}{lrrrrrrrr}
	\hline
            model &  mean &   std &  25\% &  50\% &  75\% \\
	\hline
Product           &  43.7 &  38.4 &   4.6 &  32.4 &  85.2 \\
Flatten           &  23.6 &  30.8 &   1.0 &   5.1 &  39.6 \\
Average           &  32.2 &   8.1 &  26.5 &  31.5 &  37.1 \\
Max               &   9.3 &   2.9 &   8.1 &   9.3 &  10.9 \\
LSTM              &  40.7 &  25.2 &  19.7 &  39.9 &  59.4 \\
BidirectionalLSTM &  47.8 &  25.1 &  29.6 &  50.5 &  67.7 \\
	\hline
	\end{tabular}
	\end{center}
\end{table*}

\section{Conclusion and Future Research}
In this paper was provided an insight into the possibility of using artificial neural networks for scene recognition location from a video sequence with a small set of repeated shooting locations (such as in television series) was provided. Our idea was to select more than one frame from each scene and classify the scene using that sequence of frames. We used a pretrained VGG19 network without two last layers. This results were used as an input to the trainable part our neural network architecture. We have designed six neural network models with different layer types. We have investigated different neural network layers to combine video frames, in particular average-pooling, max-pooling, product, flatten, LSTM, and bidirectional LSTM layers. The considered networks have been tested and compared on a dataset obtained from The Big Bang Theory television series. The model with max-pooling layer was not successful, its accuracy was the lowest of all models. The models with a flatten or product layer were very unstable, their standard deviation was very large. The most stable among all models was the one with an average-pooling layer. The models with unidirectional LSTM and bidirectional LSTM had similar standard deviation of the accuracy. The model with a bidirectional LSTM had the highest accuracy among all considered models. In our opinion, this is because its internal memory cells preserve information in both directions. Those results shows, that models with internal memory are able to classify with a higher accuracy than models without internal memory.

Our method may have limitations due to the chosen pretrained ANN and the low dimension of some neural layer parts. In future research, it is desirable to achieve higher accuracy in scene location recognition. This task may also need modifying model parameters or using other architectures. It also may need other pretrained models or combining several pretrained models. It is also desirable that, if the ANN detects an unknown scene, it will remember it and next time it will recognize a scene from the same location properly.

\section*{Acknowledgments}
The research reported in this paper has been supported by the Czech Science Foundation (GA\v{C}R) grant 18-18080S.

Computational resources were supplied by the project "e-Infrastruktura CZ" (e-INFRA LM2018140) provided within the program Projects of Large Research, Development and Innovations Infrastructures.

Computational resources were provided by the ELIXIR-CZ project (LM2018131), part of the international ELIXIR infrastructure.

%
%

\end{document}